\documentclass[]{fairmeta}

\usepackage{amsmath,amsfonts,bm}
\usepackage{xspace}
\usepackage{hyperref}
\usepackage{url}
\usepackage{subcaption}

\usepackage{colortbl}
\usepackage{xcolor}

\usepackage{graphicx}
\usepackage{algorithmic}
\usepackage{algorithm}
\usepackage{float}

\usepackage{booktabs}

\usepackage{enumitem}

\usepackage{amsthm}

\theoremstyle{plain}

\theoremstyle{definition}

\theoremstyle{remark}

\def\eqref#1{equation~\ref{#1}}

\def\1{\bm{1}}

\DeclareMathAlphabet{\mathsfit}{\encodingdefault}{\sfdefault}{m}{sl}
\SetMathAlphabet{\mathsfit}{bold}{\encodingdefault}{\sfdefault}{bx}{n}

\usepackage{subcaption}
\usepackage{wrapfig}
\usepackage[most]{tcolorbox}
\usepackage[most]{tcolorbox}
\usepackage{xcolor}
\usepackage{hyperref}

\definecolor{softgreen}{RGB}{110, 160, 120}

\usepackage[most]{tcolorbox} 

\newtcolorbox{takeawaybox_basemodel}[1]{
    colback=orange!5!white,    
    colframe=black,            
    arc=5pt,                   
    outer arc=5pt,
    boxrule=0.8pt,             
    left=5pt,                 
    right=5pt,                
    top=4pt,                   
    bottom=4pt,                
    fontupper=\small,          
    enhanced,
    before upper={\textbf{#1: }} 
}

\newtcolorbox{takeawaybox_rlmodel}[1]{
    colback=blue!5!white,    
    colframe=black,            
    arc=5pt,                   
    outer arc=5pt,
    boxrule=0.8pt,             
    left=5pt,                 
    right=5pt,                
    top=4pt,                   
    bottom=4pt,                
    fontupper=\small,          
    enhanced,
    before upper={\textbf{#1: }} 
}

\usepackage{caption}
\usepackage{wrapfig}    
\usepackage{booktabs}   
\definecolor{earlyblue}{HTML}{88A2F1}
\definecolor{midgrey}{HTML}{fadcb4}
\definecolor{latered}{HTML}{EE9C88}
\definecolor{highlightgreen}{HTML}{80c66d}
\definecolor{highlightpurple}{HTML}{9b6d97}
\def\thickhline{\noalign{\hrule height.8pt}}
\usepackage{arydshln} 
\usepackage{xstring}
\newcommand{\deltaval}[1]{%
  \IfBeginWith{#1}{+}{%
    {\textcolor{highlightgreen}{\textit{(#1)}}}%
  }{%
    \IfBeginWith{#1}{-}{%
      {\textcolor{highlightpurple}{\textit{(#1)}}}%
    }{%
      {\textit{(#1)}}%
    }%
  }%
}
\usepackage{pifont}

\usepackage[export]{adjustbox}

\title{What does RL improve for Visual Reasoning? A Frankenstein-Style Analysis}

\author[1,*]{Xirui Li}
\author[1,*]{Ming Li}
\author[2]{Tianyi Zhou}
\affiliation[1]{University of Maryland}
\affiliation[2]{Mohamed bin Zayed University of Artificial Intelligence}

\contribution[*]{Co-first Author}

\abstract{
Reinforcement learning (RL) with verifiable rewards has become a standard post-training stage for boosting visual reasoning in vision-language models, yet it remains unclear what capabilities RL actually improves compared with supervised fine-tuning as cold-start initialization (IN). End-to-end benchmark gains conflate multiple factors, making it difficult to attribute improvements to specific skills.
To bridge the gap, we propose a Frankenstein-style analysis framework, including: 
(i) functional localization via causal probing; 
(ii) update characterization via parameter comparison; 
and (iii) transferability test via model merging.
Instead, RL induces a consistent inference-time shift primarily in mid-to-late layers, and these mid-to-late refinements are both transferable (via merging) and necessary (via freezing) for RL gains. 
Overall, our results suggest that RL’s reliable contribution in visual reasoning is not a uniform enhancement of visual perception, but a systematic refinement of mid-to-late transformer computation that improves vision-to-reasoning alignment and reasoning performance, highlighting the limitations of benchmark-only evaluation for understanding multimodal reasoning improvements.

}

\date{\today}
\correspondence{Last authors at \email{tianyi.david.zhou@gmail.com}}

\metadata[Project Page]{\url{https://github.com/tianyi-lab/Frankenstein}\\}

\begin{document}

\maketitle

\section{Introduction}
\label{sec:introduction}
Recent progress in large language models (LLMs) post-training for reasoning tasks has been driven by a two-stage paradigm: supervised finetuning (SFT) as an initialization stage (IN) followed by reinforcement learning (RL) with verifiable rewards~\citep{guo2025deepseek, zhang2025survey, wen2025reinforcement}. 
Compared to SFT-only post-training, RL leverages sparse reward feedback and delayed credit assignment, and has been shown to elicit strong reasoning behaviors with minimal supervision in language-only settings~\citep{guo2025deepseek, zhang2025survey, wen2025reinforcement}.
Motivated by these successes, the same IN+RL\footnote{Unless otherwise specified, \textbf{Base model}, \textbf{IN model}, and \textbf{RL model}
respectively refer to the baseline, supervised initialization, and
reinforcement-learning models within a two-stage post-training pipeline.} paradigm has been increasingly adopted for vision-language models (VLMs)~\citep{zhou2025reinforced, li2025perception}, where multiple works report substantial gains on visual reasoning benchmarks across spatial reasoning~\citep{meng2025mm, huang2025vision} and multimodal understanding~\citep{deng2025openvlthinker, liu2025visual}.

Despite these gains, it remains unclear \textit{what is actually improved by RL} in visual reasoning.
Prior studies show that vision does not reliably benefit from increased inference length alone~\citep{tian2025more, rahmanzadehgervi2024vision, fu2025hidden, qin2025chain}, and verbose reasoning can even exacerbate visually grounded errors by amplifying reliance on language priors~\citep{chu2025qwen, fan2025missing, liu2025more}.
Our experiments reinforce this uncertainty by showing that the end-to-end benchmark accuracy alone cannot distinguish improvements in vision, vision-to-reasoning alignment, and pure reasoning.
At the same time, we observe the consistently increased attention from reasoning tokens to vision tokens under RL across different training recipes, suggesting a shared RL training effect whose functional role remains unclear.
These two observations motivate us to a more concise question: \textit{\textbf{What is consistently improved by RL for visual reasoning across training recipes?}}

\begin{figure*}[t]
  \centering
  \includegraphics[width=1.0\textwidth]{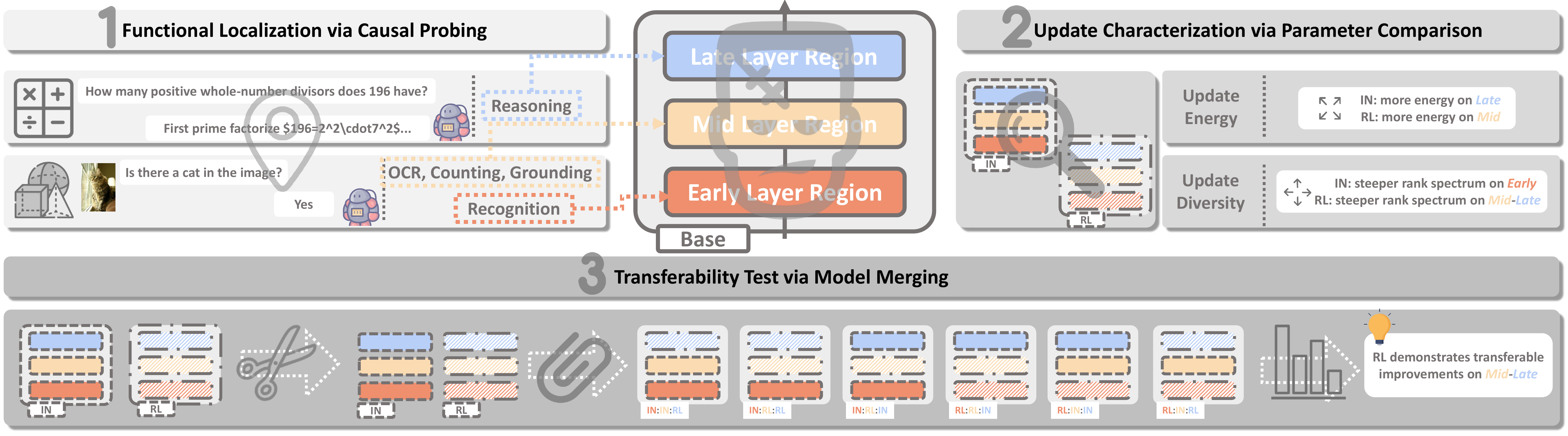}
  \caption{
\textbf{Frankenstein-style Analysis Framework.}
The framework proceeds through three components:
(1) functional localization via causal probing
across transformer depth,
(2) update characterization via parameter
comparison to identify region-wise update pattern in post-training
, and
(3) transferability test via model merging, assessing whether the localized functionalities are preserved in layers.}

  \label{fig:main_graph}
\end{figure*}

We hereby adopt a Frankenstein-style analysis framework as shown in Figure \ref{fig:main_graph} to trace \emph{where} RL alters VLMs and \emph{what} is altered across different training recipes.
By analogy to Frankenstein’s construction, we analyze VLMs by disassembling them into functional regions, intervening on these components, and reassembling them to test the causal contribution of RL-induced changes.
First, we perform \textbf{\textit{Functional Localization via Causal Probing}}, localizing vision- and reasoning-related computations along transformer depth and establishing a coarse Early/Mid/Late functional regions.
Second, we conduct \textbf{\textit{Update Characterization via Parameter Comparison}}, showing that IN and RL differ systematically in both update magnitude and update geometry, with RL exhibiting refinements concentrated in mid-late layers.
Finally, we run the \textbf{\textit{Transferability Test via Model Merging}}: transplanting RL-refined regions of layers into IN models yields consistent improvements, primarily in vision-to-reasoning alignment and reasoning, indicating that the associated functional behaviors are modular and transferable.

In addition, we further perform \textbf{\textit{Necessity Validation via Model Freezing}} during RL training.
Freezing the parameters in mid or late regions largely removes RL gains, whereas freezing those in early regions has a much weaker effect, suggesting that mid-late refinement is a critical driver of RL improvements.
Taken together, our Frankenstein-style analysis clarifies what is consistently learned under RL across training recipes and highlights the limitations of benchmark-only evaluation for understanding visual reasoning improvements in VLMs.





\textbf{Key Findings:}

1. Despite apparent gains on end-to-end visual reasoning benchmarks, fine-grained evaluation reveals that \textbf{vision ability, language-side reasoning ability, and vision-to-reasoning alignment do not improve monotonically} from the Base model to the IN model and then to the RL model.

2. Across diverse training recipes,\textbf{ RL consistently induces a shift in inference behavior, characterized by increased attention }from reasoning tokens to visual tokens, primarily in mid-late transformer layers.

3. At the parameter level, \textbf{RL exhibits a consistent and structured update pattern across recipes: refinements concentrated in mid-late layers.}
These refinements are transferable and contribute primarily to improved vision-to-reasoning alignment and reasoning performance.

\section{Motivation}
\label{sec:motivation}

\begin{figure*}[t]
  \centering
  \includegraphics[width=1.0\textwidth]{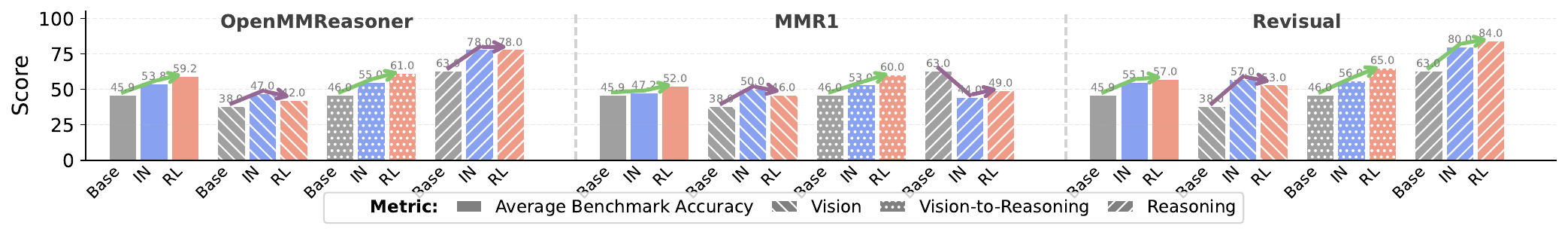}
  \caption{
\textbf{Average Benchmark Accuracy versus Fine-Grained Abilities (Vision, Reasoning, and Vision-to-Reasoning Alignment).} 
The \textcolor{highlightgreen}{\textbf{green arrows}} denote model post-training pipelines (Base Model $\rightarrow$ IN Model $\rightarrow$ RL Model) that exhibit monotonic performance gains, whereas the \textcolor{highlightpurple}{\textbf{purple arrows}} indicate model groups that do not.
Despite apparent improvements on visual reasoning benchmarks, fine-grained evaluation metrics reveal that vision ability and reasoning ability do not improve monotonically from the base model to the IN model and then to the RL model. 
}
  \label{fig:motivation_comparison}
\end{figure*}

\subsection{The Ambiguity on Visual Reasoning Improvements}
\label{sec:problem-underdetermined}

The ambiguity of RL improvements is rooted in a fundamental asymmetry between
\textit{reasoning} and \textit{vision} in current VLMs.
Reasoning ability in language models is known to improve with extended
inference, trajectory exploration, and delayed credit assignment, as
demonstrated by chain-of-thought prompting and reinforcement learning for
reasoning
\citep{wei2022chain, wang2022self, wang2025reinforcement, zhou2025r1}.

Vision, by contrast, does not reliably benefit from extended
inference alone.
Additional tokens do not introduce new visual evidence~\citep{tian2025more}, nor do they
systematically resolve perceptual errors such as mislocalized objects, missing
attributes, or incorrect counts
\citep{rahmanzadehgervi2024vision, fu2025hidden, qin2025chain}.
Moreover, extended language-side reasoning may even exacerbate visually grounded
errors by amplifying reliance on linguistic priors when visual representations
are imperfect
\citep{chu2025qwen, fan2025missing, liu2025more}.

As a result, improvements on visual reasoning benchmarks alone do not reveal whether a model’s vision or reasoning capability has actually improved.
Aggregate accuracy cannot distinguish among different sources of improvement: the same performance increase may arise from stronger reasoning over unchanged visual inputs, from changes in how reasoning attends to existing visual evidence, or from genuine improvements in visual representations.
\begin{table}[t!]
    \centering
    \small
    \caption{
        \textbf{Fine-grained evaluation metrics targeting vision, vision-to-reasoning alignment, and pure reasoning ability.}  We let $f(\cdot)$ denote the VLM’s predicted answer given its inputs. And notation $i$ refers to the original image, $b$ to a black image of the same size,
        $d$ to the textual description of the original image, and $p$ to the prompt.
    }
    \label{tab:behavior-metrics}
    \setlength{\tabcolsep}{5pt}
    \resizebox{0.5\columnwidth}{!}{
        \begin{tabular}{lll}
            \thickhline
            \toprule
            \textbf{Metric}
            & \textbf{Task}
            & \textbf{Definition} \\
            \midrule
            $M_{\text{vis}}$
            & General VQA
            & $\frac{1}{N}\sum_{n=1}^{N}\mathbb{I}\!\big[f(i_n,p_n)=y_n \,\wedge\, f(b_n,p_n)\neq y_n\big]$ \\
            \addlinespace
            $M_{\text{v2r}}$
            & Math VQA
            & $\frac{1}{N}\sum_{n=1}^{N}\mathbb{I}\!\big[f(i_n,p_n)=y_n \,\wedge\, f(b_n,d_n,p_n)=y_n\big]$ \\
            \addlinespace
            $M_{\text{rea}}$
            & Textual Math
            & $\frac{1}{N}\sum_{n=1}^{N}\mathbb{I}\!\big[f(p_n)=y_n\big]$ \\
            \bottomrule
            \thickhline
        \end{tabular}
    }
\end{table}

\subsection{Inconsistent Fine-grained Improvements}


To analyze this ambiguity, we introduce three fine-grained metrics that disentangle visual reasoning:
(i) \textit{visual perception}, (ii) \textit{vision-to-reasoning alignment}, and (iii) \textit{language-side reasoning} (Table~\ref{tab:behavior-metrics}).
This design helps separate functional gains from confounds such as hallucinated answers~\citep{fan2025missing} and spurious reliance on visual inputs~\citep{brown2025benchmark}:
\begin{itemize}
    \item \textbf{Vision ($M_{\text{vis}}$).}
    We measure whether the image provides \emph{necessary} information beyond language priors.
    Specifically, $M_{\text{vis}}$ counts an example as correct only if the model answers correctly with the real image $(i,p)$, but fails when the image is replaced by a black image $(b,p)$.
    This criterion filters out cases that can be solved without vision and focuses on instances where visual evidence is causally used.

    \item \textbf{Vision-to-Reasoning alignment ($M_{\text{v2r}}$).}
    To test whether perceptual evidence is \emph{consistently incorporated} into reasoning, we compare two semantically matched inputs:
    the original image $(i,p)$ and a textualized version of the same visual content paired with a black image $(b,d,p)$.
    $M_{\text{v2r}}$ requires the model to answer correctly in both settings, indicating that the model can preserve the correct reasoning outcome when visual evidence is replaced by an equivalent description, rather than relying on image-specific shortcuts.

    \item \textbf{Reasoning ($M_{\text{rea}}$).}
    Finally, $M_{\text{rea}}$ isolates language-side reasoning by evaluating performance on text-only problems $p$ without any visual input.
    This provides a control for improvements arising solely from strengthened reasoning.
\end{itemize}



We conduct evaluation on OpenMMReasoner~\citep{zhang2025openmmreasoner}, MMR1~\citep{leng2025mmr1}, and Revisual~\citep{chen2025advancing} families, as they include publicly available training recipes that
(1) follow the dominant RL-based post-training paradigm and (2) provide reproducible checkpoints across training stages, enabling mechanistic comparison.
For end-to-end evaluation, we also report average benchmark accuracy
across MathVista~\citep{lu2023mathvista}, MathVerse~\citep{zhang2024mathverse},
MathVision~\citep{wang2024measuring}, and LogicVista~\citep{xiao2024logicvista}.


As shown in Figure~\ref{fig:motivation_comparison}, average benchmark accuracy exhibits a monotonic increase from Base models to IN models and then to RL models.
However, these gains do not translate into consistent improvements in the fine-grained abilities isolated by our metrics:
neither vision ($M_{\text{vis}}$) nor standalone reasoning ($M_{\text{rea}}$) improves monotonically across training stages.
This discrepancy highlights a key limitation of end-to-end evaluation: they conflate qualitatively different internal changes and cannot reveal which functional components are actually strengthened by RL.

\begin{takeawaybox_basemodel}{Take-away}
Despite apparent improvements on visual reasoning benchmarks,
fine-grained evaluation reveals that vision and reasoning ability do not improve monotonically
from the Base model to the IN model and then to the RL model.
\end{takeawaybox_basemodel}

\subsection{Consistent Attention Patterns}
\label{sec:attention}

\begin{figure*}[h!]
  \centering
  \includegraphics[width=0.9\textwidth]{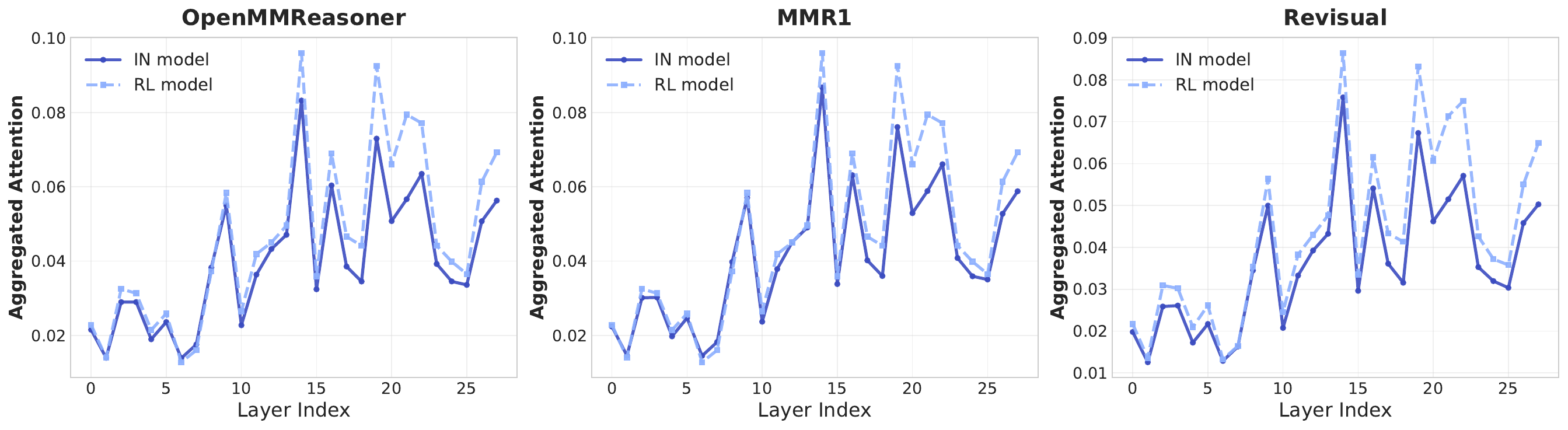}

    \caption{
    \textbf{Aggregated Attention from Reasoning Tokens to Vision Tokens.} Compared to IN models, there is more attention from reasoning tokens to vision tokens in RL models' inference.
    The pattern is concentrated in later layers across training recipes, while absent in earlier layers.
    }
    
  \label{fig:attention}
\end{figure*}


In contrast to the inconsistent improvements observed on fine-grained metrics, we identify a consistent inference-time change induced by RL: increased attention from reasoning tokens to visual tokens.
Formally, for transformer layer $\ell$ and attention head $h$, let
$A^{(\ell,h)}$ denotes the corresponding self-attention matrix, where each row sums to one.
Given a set of reasoning tokens $\mathcal{R}$ and visual tokens $\mathcal{V}$,
we define the attention mass from reasoning to vision as
\begin{equation}
\label{eq:attention-mass}
\mathcal{A}^{(\ell)}(\mathcal{R} \!\rightarrow\! \mathcal{V})
=
\frac{1}{|\mathcal{H}||\mathcal{R}||\mathcal{V}|}
\sum_{h \in \mathcal{H}}
\sum_{i \in \mathcal{R}}
\sum_{j \in \mathcal{V}}
A^{(\ell,h)}_{ij}.
\end{equation}
As shown in Figure~\ref{fig:attention}, RL modifies attention in a highly consistent manner across training recipes:
attention from reasoning tokens to visual tokens increases in mid-late layers, while attention patterns in early layers remain largely unchanged.

\begin{takeawaybox_basemodel}{Take-away}
Compared to IN models, RL models exhibit stronger attention from reasoning tokens to vision tokens.
The effect is concentrated in later layers across training recipes, while absent in earlier layers.
\end{takeawaybox_basemodel}

\section{Frankenstein-Style Analysis}
\label{sec:Frankenstein}
The contrast between inconsistent improvements on fine-grained metrics and consistent attention patterns across training recipes raises a key question:
\textit{\textbf{What is consistently improved by RL in visual reasoning across training recipes?}} We answer the question by a Frankenstein-style analysis framework that
locates RL consistent change at the \textbf{granularity of transformer layers}. 
Specifically, the framework consists of three components: (1) establishing a functional layer region partition that localizes
vision- and reasoning-related computations
(Section~\ref{sec:location}), (2) identify distinct and systematic patterns of RL updates on those layers
(Section~\ref{sec:difference}) and (3) validate the transferability of those behaviors through layer-wise model merging (Section~\ref{sec:sufficiency}).

\subsection{Functional Localization via Causal Probing}
\label{sec:location}

\begin{figure*}[h]
    \begin{subfigure}[t]{0.45\textwidth}
        \centering
        \includegraphics[width=\linewidth]{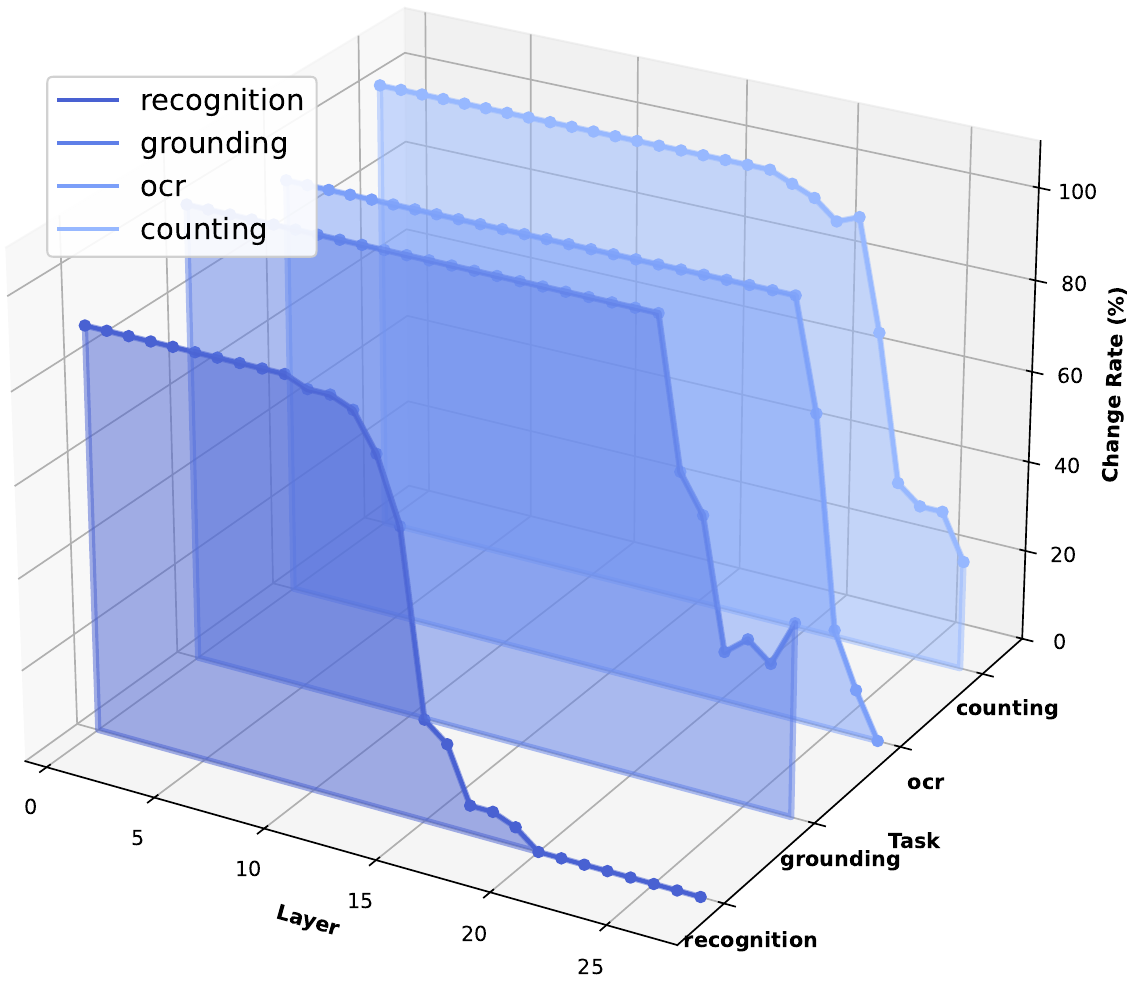}
        \caption{\textbf{Vision localization via vision-token swapping.}}
        \label{fig:visual-localization-token-swap}
    \end{subfigure}
    \centering
    \begin{subfigure}[t]{0.45\textwidth}
        \centering
        \includegraphics[width=\linewidth]{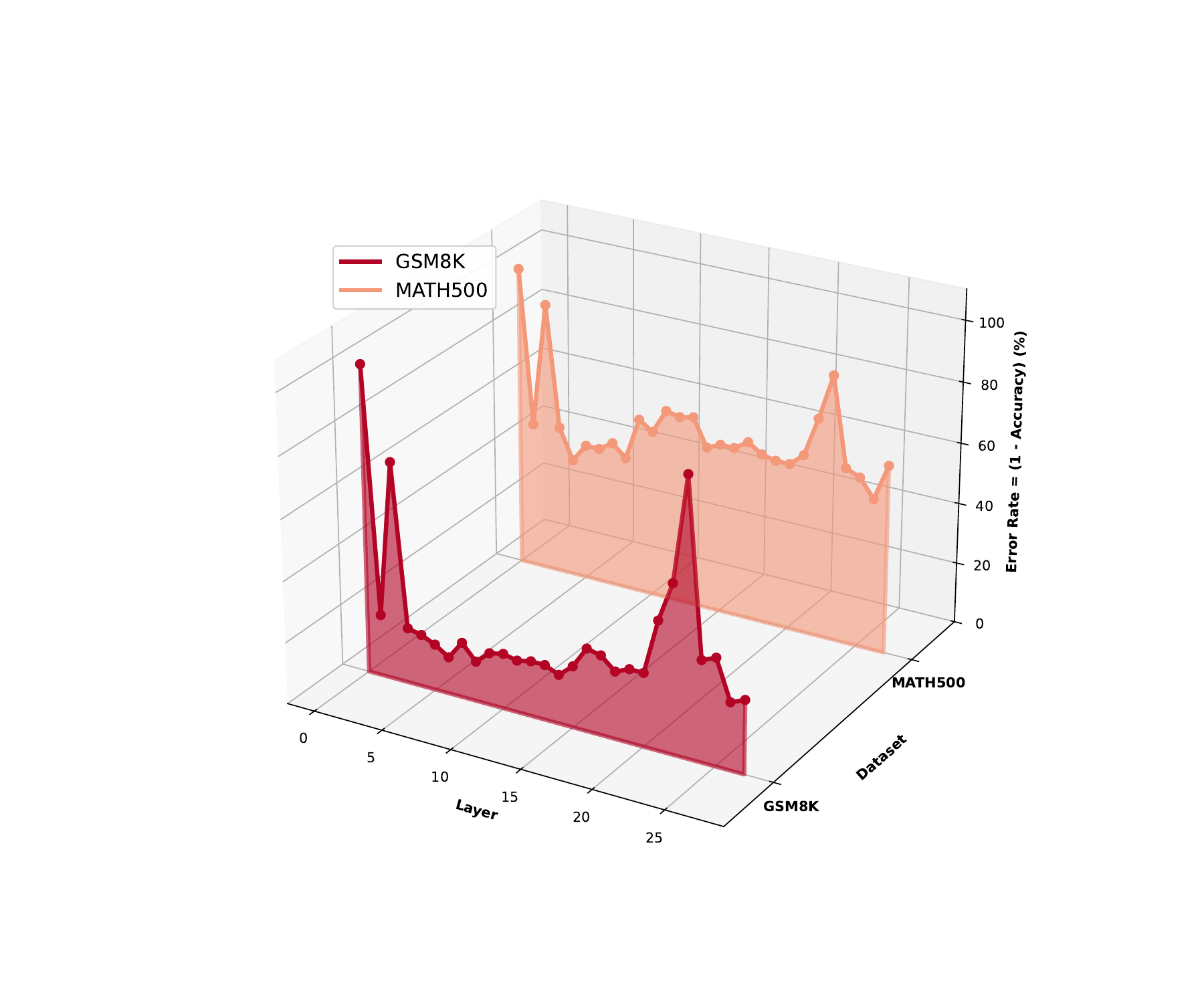}
        \caption{\textbf{Reasoning localization via layer skipping.}}
        \label{fig:reasoning-zero-ablation}
    \end{subfigure}\hfill

    \caption{\textbf{Layer-wise functional localization of vision and reasoning.}
    Both plots indicate the relative importance of each layer for the evaluated task.
    Vision-related functionalities are primarily associated with
    \emph{\textcolor{earlyblue}{\textbf{Early}}} and
    \emph{\textcolor{midgrey}{\textbf{Mid}}} transformer layers, whereas
    reasoning-related computations are concentrated in
    \emph{\textcolor{latered}{\textbf{Late}}} layers.
    }
    \label{fig:functional-localization-merged}
\end{figure*}
To understand how RL improves visual reasoning, we begin by testing whether vision and reasoning can be functionally localized to distinct regions within VLMs, enabling the subsequent region-wise analyses.
Specifically, we analyze the baseline model prior to any IN and RL training and apply
minimal, targeted interventions to localize vision- and reasoning-related
functionalities across transformer layers.
This localization establishes a layer-wise functional reference frame that
serves as the foundation for our analyses of RL training effects.

\subsubsection{Localizing vision regions}
To identify where visual information is functionally processed in the base
model, we adopt a targeted vision-token intervention strategy inspired by
\citet{shi2025vision}.
Rather than preserving in-distribution hidden states, our goal is to causally probe each layer’s contribution to the final prediction through controlled interventions.
Following the practice in \citet{shi2025vision}, we construct paired images that differ in exactly one visual
attribute (e.g., text difference in OCR task), while all other visual factors are held constant.
Concrete examples are shown in Appendix~\ref{fig:appendix_vision_demo}.
For each image pair $(i, i')$, we replace the visual tokens at a chosen
transformer layer $\ell$ with those from the paired image.
If swapping visual tokens at layer $\ell$ leads to a change in the model’s
prediction, this indicates that layer $\ell$ functionally consumes visual
information relevant to the swapped attribute.
We quantify attribute sensitivity at layer $\ell$ using the \emph{change rate},
defined as
\begin{equation}
\label{eq:vision_change_rate}
\text{Change Rate}^{(\ell)} =
\frac{1}{N}
\sum_{n=1}^{N}
\mathbb{I}
\big[
f(i_n^{(\ell)}, p_n) \neq f(i_n'^{(\ell)}, p_n)
\big],
\end{equation}
where $f(\cdot)$ denotes the model’s predicted answer, $p_n$ is the prompt, and
$i_n^{(\ell)}$ and $i_n'^{(\ell)}$ denote inputs in which visual tokens at layer
$\ell$ are sourced from the original and paired images, respectively.
Importantly, we interpret change rates comparatively across layers and visual
attributes, rather than as absolute measures of robustness.
Under this interpretation, structured peaks in the change rate indicate layers at
which the model actively processes the corresponding visual information.

 Based on the results demonstrated in Figure~\ref{fig:functional-localization-merged},
we summarize the localized functional patterns using a coarse partition of the
transformer layers into
\emph{\textcolor{earlyblue}{\textbf{Early}}},
\emph{\textcolor{midgrey}{\textbf{Mid}}}, and
\emph{\textcolor{latered}{\textbf{Late}}} layers, each comprising one third of the
layers, following prior region-wise analyses of transformer representations
(e.g.,~\citet{gurnee2025when}).
This abstraction allows us to focus on relative functionality rather than precise
layer indices.
Under this partition, vision functionality exhibits a clear stratification, consistent with prior work~\citep{chen2025bring}, suggesting
that vision and language-based reasoning can be decoupled within VLMs.
Simple vision processing, such as recognition, is concentrated primarily in the \emph{\textcolor{earlyblue}{\textbf{Early}}} layer.
Vision functionality that requires more processing, including OCR, grounding, and counting, needs contributions from \emph{\textcolor{earlyblue}{\textbf{Early}}} to \emph{\textcolor{midgrey}{\textbf{Mid}}} layers. 

\begin{figure*}[t]
  \centering
  \includegraphics[width=0.9\textwidth]{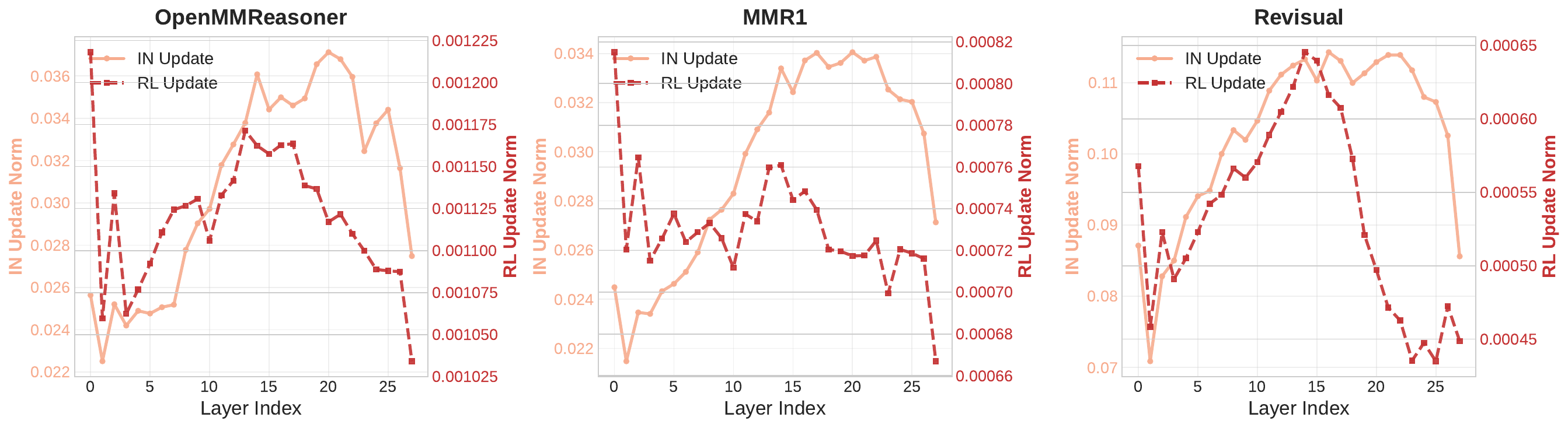}
  \caption{
  \textbf{Layer-wise parameter update norms comparison between IN and RL.}
  Per-layer Frobenius norms of parameter updates for IN (solid) and RL (dashed).
  Both training stages concentrate on optimization in the
  \emph{\textcolor{midgrey}{\textbf{Mid}}} layers, while RL exhibits a distinct
  redistribution of update magnitude compared to IN.
  }
  \label{fig:update-norm}
\end{figure*}

\subsubsection{Localizing reasoning regions}
Unlike vision, reasoning computations do not correspond to a distinct or
explicitly tagged input that can be directly manipulated.
As a result, token-level swapping is not applicable for localizing reasoning,
since the contribution of individual reasoning tokens to the final answer is
distributed and context-dependent.
We therefore employ a layer-wise causal ablation strategy, following prior
analyses of mathematical reasoning in language models~\citep{nepal2025layer}.
Rather than perturbing the input, this intervention directly removes the
contribution of a specific transformer layer while preserving the computations
of all remaining layers.
Specifically, for each transformer layer, we perform layer skipping by setting transformer layer's input as output, allowing us to assess its causal contribution to reasoning performance.
We evaluate these interventions on textual reasoning-heavy benchmarks, including mathematical reasoning tasks such as GSM8K~\citep{cobbe2021training} and MATH500~\citep{lightman2023let}.
For each ablated layer, we measure the resulting \emph{error rate}, defined as
$1 - \mathrm{Acc}$, to quantify its contribution to the reasoning process.
A substantial increase in error rate indicates
that the ablated layer plays a critical role in reasoning computation.
Additional analyses and examples of reasoning functionality localization are
provided in Appendix~\ref{app:reasoning-localization}.

In contrast to vision functionality located in earlier layers, reasoning functionality localization reveals a complementary pattern: layer-wise skipping ablation shows that the majority of the reasoning process is concentrated in the \emph{\textcolor{latered}{\textbf{Late}}} layers, with comparatively
minor dependence on early layers.

We have localized vision and reasoning functionality at the granularity of layers, but emphasize that these regions only serve as a functional
reference frame rather than precise architectural boundaries.
All subsequent analyses of decoupling and RL-based post-training effects are
interpreted with respect to this coarse \emph{\textcolor{earlyblue}{\textbf{Early}}}-\emph{\textcolor{midgrey}{\textbf{Mid}}}-\emph{\textcolor{latered}{\textbf{Late}}} functionality partition. 

\begin{takeawaybox_basemodel}{Take-away}
\emph{\textcolor{earlyblue}{\textbf{Early}}} layers primarily support simple
visual processing, \emph{\textcolor{midgrey}{\textbf{Mid}}} layers handle
higher-level visual information, and
\emph{\textcolor{latered}{\textbf{Late}}} layers are more heavily involved
in reasoning computations.
\end{takeawaybox_basemodel}

\subsection{Update Characterization via Parameter Comparison}
\label{sec:difference}

Having established a functional reference frame that localizes vision and
reasoning behaviors, we next examine whether RL updates exhibit distinct and systematic patterns within
these localized regions.
Recent studies have shown that RL can
affect reasoning behavior in ways that differ qualitatively from its
IN~\citep{zhu2025path}.
Motivated by this observation, we characterize the post-training changes through the geometry
of parameter updates, focusing on their magnitude and structure across
layers. 


\label{sec:training-difference}
For each training recipe, we compute layer-wise parameter updates
$\Delta \mathbf{W}^{(\ell)} = \mathbf{W}^{(\ell)}_{\text{trained}} -
\mathbf{W}^{(\ell)}_{\text{base}}$ for each transformer layer $\ell$.
We characterize these updates using two complementary geometric measures: update energy and diversity.

\begin{figure*}[t]
  \centering
  \includegraphics[width=\textwidth]{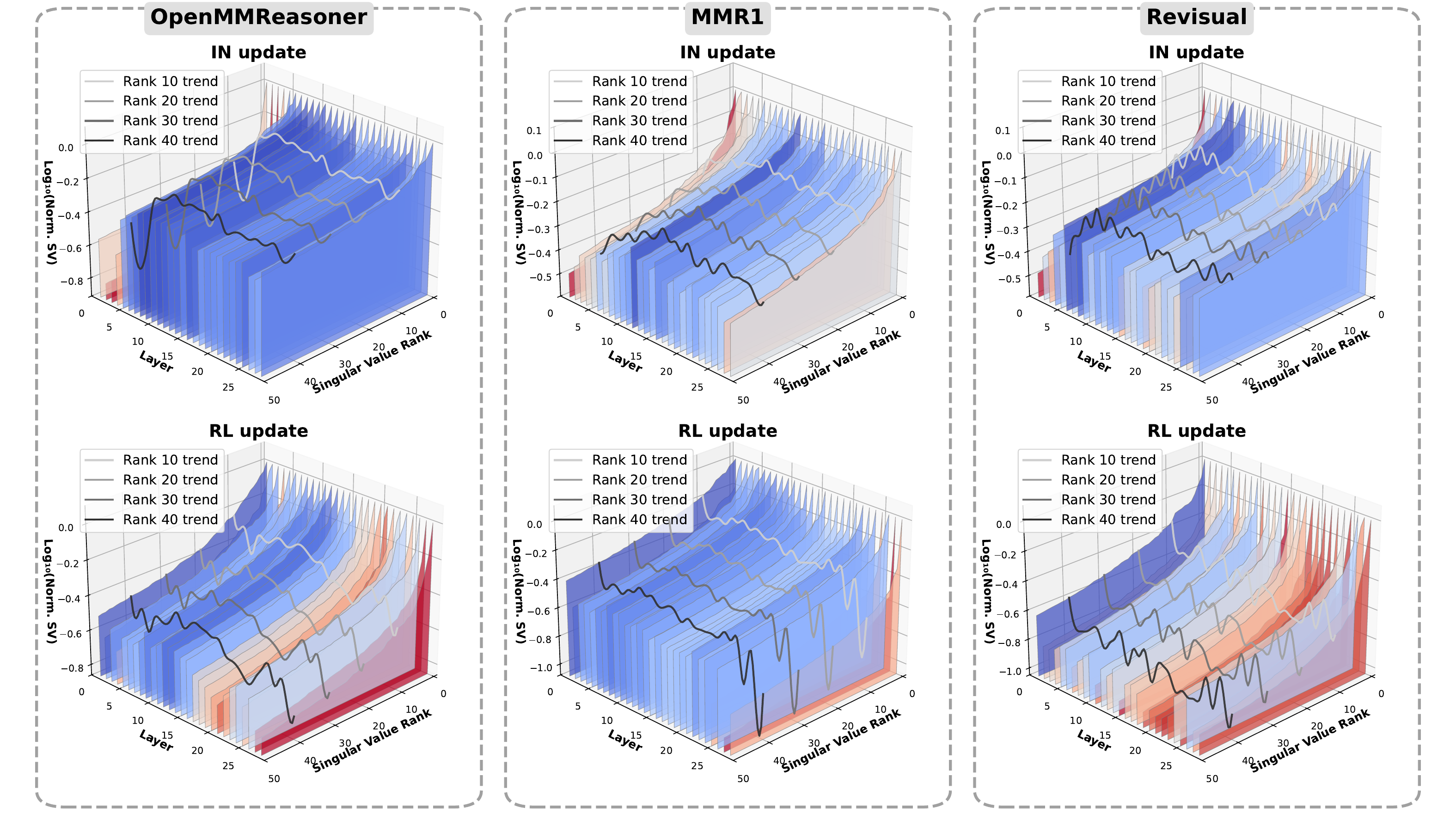}
    \caption{
    \textbf{Singular value spectra of parameter updates.}
    Each panel visualizes the
    log-normalized singular value spectrum of the layer-wise update matrix. 
    Color intensity reflects the steepness of the spectral decay.
    Compared to IN, RL updates exhibit a consistently steeper spectral decay in the
    \emph{\textcolor{midgrey}{\textbf{Mid}}}-\emph{\textcolor{latered}{\textbf{Late}}}
    layers, indicating that optimization is concentrated in a smaller number of
    dominant directions rather than being diffusely distributed.
    }
  \label{fig:update-geometry}
\end{figure*}
\subsubsection{Update Energy}
To quantify update energy, we compute the Frobenius norm
$\|\Delta \mathbf{W}^{(\ell)}\|_F$ for each layer, which measures the total
optimization magnitude accumulated during post-training.
Figure~\ref{fig:update-norm} shows that both IN and RL concentrate the majority of
their update energy in the
\emph{\textcolor{midgrey}{\textbf{Mid}}} layers, consistent with their role in
bridging visual perception and high-level reasoning.

\subsubsection{Update Diversity} To observe the update diversity, we also analyze the singular value spectrum of $\Delta \mathbf{W}^{(\ell)}$,
\begin{equation}
\label{eq:svd}
\Delta \mathbf{W}^{(\ell)} = \mathbf{U}\,\mathrm{diag}(\sigma_1,\dots,\sigma_r)\,\mathbf{V}^\top,
\end{equation}
which characterizes how optimization energy is distributed across update
directions.
For visualization, we normalize the singular values by the largest singular
value of each layer and plot their logarithm,
$\log(\sigma_i / \sigma_1)$.
This normalization maps the dominant update direction to $0$ and measures all
other directions relative to it on a log scale, removing scale differences across layers while preserving
comparative decay structure.
Under this representation, a steeper decay in the spectrum indicates that
optimization is concentrated in a small number of dominant directions,
corresponding to more focused, low-dimensional parameter refinements, whereas a
flatter spectrum reflects more diffuse updates spread across many directions.
As shown in Figure~\ref{fig:update-geometry}, each panel visualizes the
log-normalized singular value spectrum of the per-layer update matrix.
To demonstrate robustness across the spectrum, we additionally plot spectra at
every tenth rank, which exhibit consistent qualitative trends across both layers
and ranks.
Compared to IN, RL updates display a markedly steeper spectral decay in the
\emph{\textcolor{midgrey}{\textbf{Mid}}}-\emph{\textcolor{latered}{\textbf{Late}}}
layers, indicating that optimization energy is concentrated in a smaller number
of dominant directions within these regions.

\begin{takeawaybox_basemodel}{Take-away}
Both IN and RL impose high norm update in \emph{\textcolor{midgrey}{\textbf{Mid}}} layers.
However, IN and RL impose complementary optimization diversity:
RL applies less diverse refinements to
\emph{\textcolor{midgrey}{\textbf{Mid}}}-\emph{\textcolor{latered}{\textbf{Late}}} layers.
\end{takeawaybox_basemodel}

\subsection{Transferability Test via Model Merging}
\label{sec:sufficiency}
The functional localization and update analysis above establish \emph{where}
vision and reasoning are implemented and \emph{how} RL modifies different
layers of the model.
However, these analyses alone cannot determine whether the observed RL effects
reflect region-specific functional changes that only emerge in an end-to-end setting.
We hereby ask whether the improvements in vision, reasoning, and
vision-to-reasoning alignment associated with RL are \emph{transferable}, i.e.,
whether they can be preserved when parameters are updated in specific regions
transplanted into models without such training.

To answer this question, we employ \emph{model merging} as a controlled intervention.
Following the functional partition identified in Section~\ref{sec:location}, we perform model merging at the granularity of the
\emph{\textcolor{earlyblue}{\textbf{Early}}},
\emph{\textcolor{midgrey}{\textbf{Mid}}}, and
\emph{\textcolor{latered}{\textbf{Late}}} \emph{regions}.
Hybrid models are constructed by enumerating all region-wise combinations of
\emph{\textcolor{earlyblue}{\textbf{Early}}},
\emph{\textcolor{midgrey}{\textbf{Mid}}}, and
\emph{\textcolor{latered}{\textbf{Late}}} regions drawn from the IN and RL checkpoints, and by directly
copying the complete parameter state of all transformer layers within each selected region
(see Appendix~\ref{sec:appendix_model_merging} for implementation details).
This operation transfers all parameters associated with the selected transformer layers,
including any architecturally integrated vision encoder and projection layers, while leaving all other components unchanged.


Across all evaluated training recipes in Table~\ref{tab:main}, model
merging reveals a consistent and interpretable pattern in how RL-induced
behaviors transfer across models.
Hybrid models that preserve RL-refined layers in the
\emph{\textcolor{midgrey}{\textbf{Mid}}}-\emph{\textcolor{latered}{\textbf{Late}}}
layers (especially merged model
\textbf{\textcolor{earlyblue}{IN}:\textcolor{midgrey}{RL}:\textcolor{latered}{RL}})
consistently retain vision-to-reasoning alignment and reasoning improvements relative to the
\textbf{IN} model.
In contrast, alternative region combinations do not exhibit consistent or uniform
improvements across training recipes, and several configurations lead to degraded
performance.

Together, these results indicate that RL-induced refinements in the
\emph{\textcolor{midgrey}{\textbf{Mid}}}-\emph{\textcolor{latered}{\textbf{Late}}}
layers encode transferable functional behaviors that can be preserved
through model merging.
Notably, the retained gains primarily manifest as improved vision-to-reasoning
alignment and reasoning,
consistent with the 
observation of concentrated refinements in
\emph{\textcolor{midgrey}{\textbf{Mid}}}-\emph{\textcolor{latered}{\textbf{Late}}}
layers discussed in Section~\ref{sec:difference}.

\begin{table*}[t]
    \rowcolors{2}{gray!11}{white}
    \centering
    \small

    \caption{
    \textbf{Frankenstein-style model merging.}
    We report vision, reasoning, and vision-to-reasoning alignment scores for
    Base, IN, and RL checkpoints, as well as hybrid models constructed by
    transferring \emph{training-induced changes} across
    \emph{\textcolor{earlyblue}{\textbf{Early}}},
    \emph{\textcolor{midgrey}{\textbf{Mid}}},
    and \emph{\textcolor{latered}{\textbf{Late}}} transformer layers.
    We denote the performance gap between merged models and IN with \textbf{\textcolor{highlightgreen}{green on positive values}} and \textbf{\textcolor{highlightpurple}{purple on the rest}}.
    Across training recipes, when \textbf{\textcolor{midgrey}{RL}} and \textbf{\textcolor{latered}{RL}} are kept, merged models demonstrate a consistent pattern of performance improvement, indicating RL consistent refinements in the \emph{\textcolor{midgrey}{\textbf{Mid}}}-\emph{\textcolor{latered}{\textbf{Late}}} layers.}
    \label{tab:main}

    \resizebox{\textwidth}{!}{
        \begin{tabular}{
            >{\raggedright\arraybackslash}p{3.8cm} |
            *{1}{>{\centering\arraybackslash}p{1.4cm}} :
            *{1}{>{\centering\arraybackslash}p{1.4cm}} :
            *{6}{>{\centering\arraybackslash}p{2.2cm}} :
            >{\centering\arraybackslash}p{1.4cm}
        }
        \thickhline
        \toprule
        & \multicolumn{2}{c}{}
        & \multicolumn{6}{c}{\texttt{Region-wise Merged Models}}
        & \multicolumn{1}{c}{} \\
        \cmidrule(lr){4-9}
        \textbf{Ability}
            & \textbf{Base}
            & \textbf{IN}
            & \textbf{\textcolor{earlyblue}{IN} : \textcolor{midgrey}{RL} : \textcolor{latered}{RL}}
            & \textbf{\textcolor{earlyblue}{IN} : \textcolor{midgrey}{IN} : \textcolor{latered}{RL}}
            & \textbf{\textcolor{earlyblue}{RL} : \textcolor{midgrey}{RL} : \textcolor{latered}{IN}}
            & \textbf{\textcolor{earlyblue}{RL} : \textcolor{midgrey}{IN} : \textcolor{latered}{IN}}
            & \textbf{\textcolor{earlyblue}{IN} : \textcolor{midgrey}{RL} : \textcolor{latered}{IN}}
            & \textbf{\textcolor{earlyblue}{RL} : \textcolor{midgrey}{IN} : \textcolor{latered}{RL}}
            & \textbf{RL} \\
        \midrule


        \hiderowcolors
        \multicolumn{10}{c}{
            \textcolor{gray}{\textit{Training Recipe -- OpenMMReasoner~\citep{zhang2025openmmreasoner}}}
        } \\
        \showrowcolors

Vision
& 38.0 & 47.0
& 42.0 \deltaval{-5.0}
& 48.0 \deltaval{+1.0}
& 45.0 \deltaval{-2.0}
& 45.0 \deltaval{-2.0}
& 44.0 \deltaval{-3.0}
& 40.0 \deltaval{-7.0}
& 42.0 \\

Vision-to-Reasoning
& 46.0 & 55.0
& 59.0 \deltaval{+4.0}
& 58.0 \deltaval{+3.0}
& 63.0 \deltaval{+8.0}
& 57.0 \deltaval{+2.0}
& 58.0 \deltaval{+4.0}
& 55.0 \deltaval{-0.0}
& 61.0 \\

Reasoning
& 63.0 & 78.0
& 81.0 \deltaval{+3.0}
& 73.0 \deltaval{-5.0}
& 78.0 \deltaval{-0.0}
& 73.0 \deltaval{-5.0}
& 80.0 \deltaval{+2.0}
& 73.0 \deltaval{-5.0}
& 78.0 \\


\midrule
\hiderowcolors
\multicolumn{10}{c}{
\textcolor{gray}{\textit{Training Recipe -- MMR1~\cite{leng2025mmr1}}}
} \\
\showrowcolors

Vision
& 38.0 & 44.0
& 37.0 \deltaval{-7.0}
& 41.0 \deltaval{-3.0}
& 45.0 \deltaval{+1.0}
& 45.0 \deltaval{+1.0}
& 41.0 \deltaval{-3.0}
& 43.0 \deltaval{-1.0}
& 50.0 \\

V-to-R
& 46.0 & 46.0
& 51.0 \deltaval{+5.0}
& 50.0 \deltaval{+4.0}
& 50.0 \deltaval{+4.0}
& 56.0 \deltaval{+10.0}
& 52.0 \deltaval{+6.0}
& 57.0 \deltaval{+13.0}
& 54.0 \\

Reasoning
& 63.0 & 60.0
& 61.0 \deltaval{+1.0}
& 64.0 \deltaval{+4.0}
& 64.0 \deltaval{+4.0}
& 79.0 \deltaval{+19.0}
& 56.0 \deltaval{-4.0}
& 72.0 \deltaval{+12.0}
& 62.0 \\


\midrule
\hiderowcolors
\multicolumn{10}{c}{
\textcolor{gray}{\textit{Training Recipe -- Revisual~\cite{chen2025advancing}}}
} \\
\showrowcolors

Vision
& 38.0 & 40.0
& 42.0 \deltaval{+2.0}
& 42.0 \deltaval{+2.0}
& 41.0 \deltaval{+1.0}
& 37.0 \deltaval{-3.0}
& 40.0 \deltaval{-0.0}
& 46.0 \deltaval{+6.0}
& 35.0 \\

Vision-to-Reasoning
& 46.0 & 56.0
& 59.0 \deltaval{+3.0}
& 58.0 \deltaval{+2.0}
& 58.0 \deltaval{+2.0}
& 56.0 \deltaval{-0.0}
& 60.0 \deltaval{+4.0}
& 51.0 \deltaval{-5.0}
& 59.0 \\

Reasoning
& 63.0 & 81.0
& 85.0 \deltaval{+4.0}
& 84.0 \deltaval{+3.0}
& 84.0 \deltaval{+3.0}
& 85.0 \deltaval{+4.0}
& 86.0 \deltaval{+5.0}
& 88.0 \deltaval{+7.0}
& 88.0 \\

        \bottomrule
        \thickhline
        \end{tabular}}
\end{table*}

\begin{takeawaybox_basemodel}{Take-away}
Model merging confirms consistent RL refinements in \emph{\textcolor{midgrey}{\textbf{Mid}}}-\emph{\textcolor{latered}{\textbf{Late}}} layers, which consistently improve vision-to-reasoning alignment and reasoning capabilities.
\end{takeawaybox_basemodel}

\subsection{Summary of Region-wise RL Effects}
Through our proposed Frankenstein-style framework, we find that RL-induced changes in VLMs are localized
to the \emph{\textcolor{midgrey}{\textbf{Mid}}}-\emph{\textcolor{latered}{\textbf{Late}}}
transformer layers and exhibit concentrated refinements.
Region-wise model merging further shows that preserving these RL-refined layers
consistently retains improvements in vision-to-reasoning alignment and reasoning across
training recipes.
Together, these results across training recipes indicate that RLs consistently contribute to \emph{\textcolor{midgrey}{\textbf{Mid}}}-\emph{\textcolor{latered}{\textbf{Late}}} layers in vision-to-reasoning alignment and reasoning behaviors.

\section{Necessity Validation via
Model Freezing}
\label{sec:necessity}

The Frankenstein-style analysis in Section~\ref{sec:Frankenstein} shows that across different training recipes, RL consistently demonstrates concentrated  refinements on \emph{\textcolor{midgrey}{\textbf{Mid}}}-\emph{\textcolor{latered}{\textbf{Late}}} layers on vision-to-reasoning and reasoning capabilities.
However, region-wise transferability alone does not establish whether refinement
of these layers is required for RL gains to emerge during training.

To validate the region-wise RL effects identified above, we perform targeted training-time interventions via
region-wise parameter freezing during RL training (see Appendix~\ref{sec:appendix_model_training} for training details).
Following the partition identified in
Section~\ref{sec:location}, we freeze one region of layers at a time while allowing the remaining to be optimized, keeping all other training settings identical.
This design enables a necessary-condition test: if RL improvements are causally
mediated by refinement on \emph{\textcolor{midgrey}{\textbf{Mid}}}-\emph{\textcolor{latered}{\textbf{Late}}} layers, then preventing RL from updating the
\emph{\textcolor{midgrey}{\textbf{Mid}}}-\emph{\textcolor{latered}{\textbf{Late}}} layers should mitigate these gains, whereas
constraining earlier layers should have a markedly smaller effect.

As demonstrated in Table~\ref{tab:training_freeze}, freezing the \emph{\textcolor{latered}{\textbf{Late}}} layers during
RL training results in substantial degradation across both fine-grained and benchmark metrics.
In contrast, freezing the \emph{\textcolor{earlyblue}{\textbf{Early}}} and leaving \emph{\textcolor{midgrey}{\textbf{Mid}}}-\emph{\textcolor{latered}{\textbf{Late}}} layers intact yields better performance than the rest of the models, including \textbf{IN} and \textbf{RL}.
These results demonstrate that \emph{\textcolor{midgrey}{\textbf{Mid}}}-\emph{\textcolor{latered}{\textbf{Late}}} refinement is not merely correlated
with RL improvements, but it is a necessary component for achieving them. 

\begin{takeawaybox_basemodel}{Take-away}
RL gains depend causally on \emph{\textcolor{midgrey}{\textbf{Mid}}}-\emph{\textcolor{latered}{\textbf{Late}}} refinement:
Preventing RL from updating the \emph{\textcolor{latered}{\textbf{Late}}} region during training eliminates improvements in both fine-grained and benchmark metrics.
\end{takeawaybox_basemodel}

\begin{table*}[t]
    \rowcolors{2}{gray!11}{white}
    \centering
    \small

    \caption{
    \textbf{Parameter freezing during RL training.}
    We compare fine-grained and benchmark metrics for standard RL training and RL
    training with parameters in one transformer region frozen.
    Freezing the \emph{\textcolor{latered}{\textbf{Late}}} layers leads to a pronounced drop in
    reasoning performance and vision-to-reasoning alignment, whereas freezing
    earlier layers has a smaller effect.
    }
    \label{tab:training_freeze}

    \resizebox{0.9\textwidth}{!}{
        \begin{tabular}{l|ccc|ccc}
            \thickhline
            \toprule
             & \multicolumn{3}{c}{\texttt{Fine-grained Metrics}}
            & \multicolumn{3}{c}{\texttt{Benchmark Metrics}} \\
            \cmidrule(lr){2-4} \cmidrule(lr){5-7}
                & \textbf{Vision ($M_{\text{vis}}$)}
                & \textbf{Vision-to-Reasoning ($M_{\text{v2r}}$)}
                & \textbf{Reasoning ($M_{\text{rea}}$)}
                & \textbf{MathVista}
                & \textbf{MathVision}
                & \textbf{MathVerse} \\
            \midrule


            \hiderowcolors \multicolumn{7}{c}{
                \textcolor{gray}{\textit{Original Training Recipe}}
            } \\ \showrowcolors
            IN Model~\citep{zhang2025openmmreasoner}
                & 34.0 & 21.0 & 26.0 & 46.5 & 18.4 & 37.0 \\
            RL Model~\citep{zhang2025openmmreasoner}
                & 33.0 & 29.0 & 34.0 & 48.1 & 14.1 & 37.8 \\
            \midrule

            \hiderowcolors \multicolumn{7}{c}{
                \textcolor{gray}{\textit{Training with Frozen Blocks}}
            } \\ \showrowcolors
            RL Model - Frozen \emph{\textcolor{earlyblue}{\textbf{Early}}} Block
                & \textbf{35.0} & \textbf{31.0} & 36.0 & \textbf{48.2} & \textbf{21.0} & 34.5 \\
            RL Model - Frozen \emph{\textcolor{midgrey}{\textbf{Mid}}} Block
                & 25.0 & 29.0 & \textbf{38.0} & 46.5 & 15.5 & \textbf{35.7} \\
            RL Model - Frozen \emph{\textcolor{latered}{\textbf{Late}}} Block
                & 30.0 & 27.0 & 34.0 & 47.9 & 16.8 & 35.0  \\
            \bottomrule
            \thickhline
        \end{tabular}
    }
\end{table*}

\section{Related Work}

\subsection{RL and Its Analysis}

RL has emerged as a powerful post-training paradigm for improving the reasoning capabilities of large language models.
GRPO, introduced in DeepSeekMath~\citep{shao2024deepseekmath} and later scaled in DeepSeek-R1~\citep{guo2025deepseek}, forms the basics of many recent advances.
Subsequent variants, including DAPO~\citep{yu2025dapo}, GSPO~\citep{zheng2025group}, GFPO~\citep{shrivastava2025sample}, and GDPO~\citep{liu2026gdpo}, further refine the stability and efficiency of RL training.

Most prior analyses of RL focus on end-to-end performance metrics such as pass@k or accuracy improvements~\citep{yue2025does, wen2025reinforcement}, treating the model as a black box.
In contrast, only a limited number of works~\citep{mukherjee2025reinforcement, zhu2025path} investigate RL from a parameter-space perspective, examining where updates are localized and how their geometric structure differs from supervised finetuning.
Our work builds on this emerging line of inquiry by explicitly analyzing the location and geometry of RL updates in VLMs.

\subsection{Improvement of Visual Reasoning in VLMs}

Early work demonstrated that visual reasoning in VLMs can be enhanced through Chain-of-Thought finetuning~\citep{zhang2023multimodal, wei2022chain}.
More recent R1-style training pipelines show that such reasoning behaviors can be further amplified using RL; we refer readers to the Appendix~\ref{sec:appendix_more_related_work} for a comprehensive discussion of this line of work.
Alternative approaches achieve similar gains by model composition~\citep{chen2025bring}, vision tokens~\citep{bigverdi2025perception}, explicit grounding~\citep{sarch2025grounded, zhang2025perceptual}, or multi-agent framework~\citep{jia2025decoupling}.

In parallel, several studies examine how post-training affects visual processing itself.
Notably, \citet{song2025rl} shows that RL can improve the vision encoder in VLMs more effectively than SFT, suggesting that RL updates may not be confined to purely linguistic or reasoning components.

\section{Conclusion}

In this work, we address the ambiguity underlying RL-based improvements in visual reasoning by asking what is \emph{consistently} changed by RL across training recipes.
Rather than relying solely on end-to-end benchmarks, we adopt a Frankenstein-style analysis framework that decomposes VLMs at the granularity of transformer layers and probes their functional roles.
Through causal localization, update characterization, and region-wise model merging, we show that RL does not uniformly improve visual perception or standalone reasoning.
Instead, RL is consistently associated with structured refinements in middle and late transformer layers, reflected behaviorally in improved vision-to-reasoning alignment and reasoning capabilities.
This perspective clarifies how RL alters VLM behavior beyond what aggregate accuracy reveals and provides a principled framework for diagnosing improvements in visual reasoning across training recipes.

\clearpage
\newpage
\bibliographystyle{assets/plainnat}
\bibliography{main}

\clearpage
\newpage
\beginappendix

\section{Limitations}

Our study has several limitations that point to directions for future work.
First, we focus on the dominant IN+RL-style post-training paradigm rather than
direct RL from scratch.
While representative of current practice, our findings may not directly
generalize to alternative RL training recipes.
Second, our analysis is limited to VLMs aim to improve visual reasoning.
The functional decomposition identified here may not transfer to other task domains or modalities.
Finally, all experiments are conducted on models from the Qwen series~\citep{bai2025qwen2} (to the best of our knowledge, most training recipes are all based on the Qwen series). 
Although the proposed framework is model-agnostic in principle, its applicability to models without clearly separable functional regions remains to be explored.
\begin{table*}[h]
\centering
\small
\caption{\textbf{Training Recipes and Checkpoint Availability of RL-Based Visual Reasoning Models.} We compare recently proposed RL-based post-training methods along key dimensions,
including base model, training pipeline, reinforcement learning algorithm, training domain,
claimed ability gains, and checkpoint availability.
Although many methods share highly similar training recipes, only a small subset of these works sufficiently
full-fill checkpoints to support controlled, stage-wise and block-level analysis.
We therefore focus on the gray-highlighted models in this work, which follow the dominant
RL training recipe while enabling reproducible and mechanistic comparison across training stages.}
\label{tab:rlvr-recipes}

\resizebox{\textwidth}{!}{
\begin{tabular}{l l l l l l l c}
\thickhline
\toprule
\textbf{Name}
& \textbf{Date}
& \textbf{Base Model} 
& \textbf{Pipeline}
& \textbf{RL Algorithm} 
& \textbf{Training Domain} 
& \textbf{Claimed Gains} 
& \textbf{All CKPT Availability} \\
\midrule
Visual-RFT~\citep{liu2025visual}
& 25.03
& Qwen-VL
& IN + RL
& GRPO
& Vision Task 
& Vision
& \ding{56} \\

MM-Eureka~\citep{meng2025mm}
& 25.03
& Intern-VL
& RL
& GRPO 
& Math 
& Visual Reasoning 
& \ding{56} \\

Vision-R1~\citep{huang2025vision}
& 25.03
& Qwen-VL
& IN + RL 
& GRPO 
& Math 
& Visual Reasoning  
& \ding{56} \\

VisualThinker~\citep{zhou2025r1}
& 25.03
& Qwen-VL
& IN + RL 
& GRPO 
& Vision Task 
& Vision 
& \ding{56} \\

Curr-ReFT~\citep{deng2025boosting}
& 25.03
& Qwen-VL
& IN + RL 
& GRPO 
& Vision Task + Math
& Visual Reasoning 
& \ding{56} \\

VisRL~\citep{chen2025visrl}
& 25.03
& Qwen-VL, LLaVA 
& IN + RL 
& DPO 
& Vision Task 
& Vision 
& \ding{56} \\

LMM-R1~\citep{peng2025lmm}
& 25.03
& Qwen-VL
& RL 
& PPO 
& Math 
& Visual Reasoning  
& \ding{56} \\

R1-Onevision~\citep{yang2025r1}
& 25.03
& Qwen-VL
& IN + RL 
& GRPO 
& Math 
& Visual Reasoning  
& \ding{56} \\

Skywork R1V~\citep{wang2025skywork}
& 25.03
& QwQ
& RL 
& MPO + GRPO 
& Math 
& Visual Reasoning  
& \ding{56} \\

R1-VL~\citep{zhang2025r1}
& 25.03
& Qwen-VL
& RL 
& GRPO 
& Math 
& Visual Reasoning  
& \ding{56} \\

OThink-MR1~\citep{liu2025othink}
& 25.03
& Qwen-VL
& RL 
& GRPO-D
& Vision Task + Math
& Visual Reasoning  
& \ding{56} \\

OpenVLThinker~\citep{deng2025openvlthinker}
& 25.03
& Qwen-VL
& IN + RL (iterative) 
& GRPO
& Vision Task + Math
& Visual Reasoning  
& \ding{56} \\

Reason-RFT~\citep{tan2025reason}
& 25.03
& Qwen-VL
& IN + RL 
& GRPO
& Vision Task
& Vision  
& \ding{52} \\

CrowdVLM-R1~\citep{wang2025crowdvlm}
& 25.04
& Qwen-VL
& RL 
& GRPO
& Vision Task
& Vision  
& \ding{56} \\

VLM-R1~\citep{shen2025vlm}
& 25.04
& Qwen-VL
& IN + RL 
& GRPO
& Vision Task
& Vision  
& \ding{56} \\

Perception-R1~\citep{yu2025perception}
& 25.04
& Qwen-VL
& RL 
& GRPO
& Vision Task
& Vision  
& \ding{52} \\

VLAA-Thinker~\citep{yu2025perception}
& 25.04
& Qwen-VL
& IN + RL 
& GRPO
& Vision Task + Math
& Visual Reasoning
& \ding{56} \\

VL-Rethinker~\citep{wang2025vl}
& 25.04
& Qwen-VL
& IN + RL 
& GRPO
& Vision Task + Math
& Visual Reasoning
& \ding{56} \\

G1~\citep{chen2025g1}
& 25.05
& Qwen-VL
& RL 
& GRPO
& Vision Task
& Vision 
& \ding{56} \\

UniVG-R1~\citep{bai2025univg}
& 25.05
& Qwen-VL
& RL 
& IN + GRPO
& Vision Task
& Vision 
& \ding{56} \\

STAR-R1~\citep{li2025star}
& 25.05
& Qwen-VL
& RL 
& GRPO
& Vision Task
& Vision 
& \ding{56} \\

Pixel Reasoner~\citep{wang2025pixel}
& 25.05
& Qwen-VL
& IN + RL 
& GRPO
& Vision Task
& Vision
& \ding{52} \\

R1-ShareVL~\citep{yao2025r1}
& 25.05
& Qwen-VL
& RL 
& Share-GRPO
& Vision Task + Math
& Visual Reasoning
& \ding{56} \\

SRPO~\citep{wan2025srpo}
& 25.06
& Qwen-VL
& IN + RL 
& SRPO
& Vision Task + Math
& Visual Reasoning
& \ding{56} \\

Rex-Thinker~\citep{jiang2025rex}
& 25.06
& Qwen-VL
& IN + RL 
& GRPO
& Vision Task
& Vision
& \ding{56} \\

\rowcolor{gray!15}
ReVisual-R1~\citep{chen2025advancing}
& 25.06
& Qwen-VL
& IN + RL 
& GRPO
& Vision Task + Math
& Visual Reasoning
& \ding{52} \\

APO~\citep{hong2025apo}
& 25.06
& Qwen-VL
& RL 
& APO
& Vision Task + Math
& Visual Reasoning
& \ding{52} \\

PAPO~\citep{wang2025perception}
& 25.07
& Qwen-VL
& RL 
& PAPO
& Vision Task + Math
& Visual Reasoning
& \ding{52} \\

\rowcolor{gray!15}
MMR1~\citep{leng2025mmr1}
& 25.09
& Qwen-VL
& IN + RL 
& GRPO
& Vision Task + Math
& Visual Reasoning
& \ding{52} \\

\rowcolor{gray!15}
OpenMMReasoner~\citep{zhang2025openmmreasoner}
& 25.11
& Qwen-VL
& IN + RL 
& GRPO
& Vision Task + Math
& Visual Reasoning
& \ding{52} \\

\bottomrule
\thickhline
\end{tabular}
}
\end{table*}
\section{More Related Work}
\label{sec:appendix_more_related_work}

\subsection{RL-based Visual Reasoning Improvements}

Recent work has proposed a large number of RL-based post-training  methods for
improving visual reasoning in VLMs. As summarized in
Table~\ref{tab:rlvr-recipes}, however, this apparent diversity masks a highly concentrated
design space. Most methods adopt the same underlying base model, follow
a similar two-stage pipeline consisting of instruction tuning followed by reinforcement
learning (IN\,+\,RL), and rely on closely related policy optimization algorithms such as
GRPO. 
Differences across methods are therefore largely confined to task mixtures, reward designs, or training heuristics, rather than to fundamentally distinct training paradigms.

For the purposes of this work, our goal is not to compare end-to-end benchmark performance
across all proposed methods, but to understand \emph{what internal changes are induced by
RL-based post-training}. 
Achieving this requires controlling for the training recipe itself:
we fix a common and widely adopted training pipeline so that observed differences can be
attributed to RL-induced updates rather than to heterogeneous optimization schemes or
architectural choices. In addition, such analysis fundamentally depends on access to
complete and well-aligned checkpoints across training stages, which are necessary for
stage-wise comparison and layer-level interventions.

We therefore focus on three representative models that are highlighted in Table~\ref{tab:rlvr-recipes}: ReVisual-R1, MMR1, and
OpenMMReasoner. 
First, they follow the dominant RL training
recipe and target visual reasoning using mixed vision and math supervision, making them
representative of current practice. 
Second, they provide sufficiently
complete checkpoint releases to enable reproducible, stage-wise comparison. 
By restricting
our analysis to these models, we can isolate RL-induced effects while avoiding confounding
factors arising from architectural differences or incomplete training artifacts.

\section{Fine-grained Metrics}
\label{app:finegrained_metrics}

This appendix details the fine-grained evaluation metrics used in
Table~\ref{tab:behavior-metrics}.
We decompose visual reasoning into three abilities: \emph{Vision},
\emph{Vision-to-Reasoning Alignment}, and \emph{Reasoning}, and define each metric using
paired input settings that selectively control the availability of visual
evidence.

\subsection{Taxonomy}

We decompose visual reasoning into three complementary abilities that capture
distinct aspects of multimodal behavior.

\paragraph{Vision.}
\emph{Vision} measures whether a model can extract task-relevant information from
visual input in a way that \emph{changes} the final prediction.
A model with strong vision should produce correct answers in cases where
language-only inference fails, indicating that visual evidence contributes
non-trivially to decision making.

\paragraph{Vision-to-Reasoning.}
\emph{Vision-to-Reasoning} measures whether perceptual evidence is
\emph{consistently and correctly} incorporated into downstream reasoning.
This ability captures alignment between visual perception and reasoning by
assessing whether the model arrives at correct answers under both real-image
input and textualized visual input representing the same underlying information.

\paragraph{Reasoning.}
\emph{Reasoning} measures language-side inference ability independent of visual
input.
It reflects the model’s capacity for multi-step
reasoning when no visual evidence is available.

\subsection{Metric Definitions}

We use the following notation, consistent with
Table~\ref{tab:behavior-metrics}.
Let $i$ denote the original image, $b$ a black image of the same size, $d$ a
textual description of the original image, and $p$ the task prompt.
Let $f(\cdot)$ denote the model prediction and $y$ the ground-truth answer.
All metrics are computed over $N$ evaluation samples.

\paragraph{Vision.}
Vision is evaluated on General VQA tasks using an unconditional paired metric:
\begin{equation}
M_{\text{vis}}
=
\frac{1}{N}
\sum_{n=1}^{N}
\mathbb{I}\!\left[
f(i_n, p_n)=y_n
\;\wedge\;
f(b_n, p_n)\neq y_n
\right].
\end{equation}
This metric measures the fraction of instances whose correct prediction is
\emph{enabled by visual input}.
Because it uses the full evaluation set as the denominator, VisionScore supports
direct comparison across checkpoints.

\paragraph{Vision-to-Reasoning.}
Vision-to-Reasoning is evaluated on Math VQA tasks using a joint correctness
criterion:
\begin{equation}
M_{\text{v2r}}
=
\frac{1}{N}
\sum_{n=1}^{N}
\mathbb{I}\!\left[
f(i_n, p_n)=y_n
\;\wedge\;
f(b_n, d_n, p_n)=y_n
\right].
\end{equation}
Here, $b, d, p$ replaces the image with a textual description of the same visual
content.
A higher score indicates stronger vision-to-reasoning alignment, as the model
produces correct and consistent reasoning outcomes under both visual and
textualized-visual inputs.
Using the full dataset as the denominator avoids model-dependent conditional
subsets and enables cross-checkpoint comparison.

\paragraph{Reasoning.}
Reasoning is evaluated on textual math tasks as:
\begin{equation}
M_{\text{rea}}
=
\frac{1}{N}
\sum_{n=1}^{N}
\mathbb{I}\!\left[
f(p_n)=y_n
\right].
\end{equation}
This setting isolates
language-side reasoning ability.

\subsection{Evaluation Tasks and Datasets}

We instantiate the above metrics using the following datasets:

\begin{itemize}
    \item \textbf{Vision (General VQA).}
    AI4Math/MathVista~\citep{lu2023mathvista} \texttt{testmini} split with
    \texttt{metadata['category'] = general-vqa},

    \item \textbf{Vision-to-Reasoning (Math VQA).}
    AI4Math/MathVista~\citep{lu2023mathvista} \texttt{testmini} split with
    \texttt{metadata['category'] = math-targeted-vqa},

    \item \textbf{Reasoning (Textual Math).}
    HuggingFaceH4/MATH-500~\citep{lightman2023let} \texttt{test} split.
\end{itemize}

\section{Functional Region Localization}
\label{app:functional-localization}

This appendix provides full experimental details for the functional region
localization experiments described in Section~\ref{sec:difference}.
These experiments are designed as \emph{causal probes} to identify where visual
and reasoning signals are functionally consumed in the base model.
They serve to establish a functional reference frame for interpreting the
effects of RL-based post-training.

\subsection{Design Principles}

All functional localization experiments follow a shared design principle:
\textbf{minimal, localized intervention under fixed inference-time computation}.
Any observed output change is therefore attributable to the targeted
intervention rather than differences in inference depth or search.

All experiments are conducted on the \emph{base} checkpoint (in our case \textbf{Qwen2.5-VL-7B-Instruct}~\citep{bai2025qwen2}) to ensure
that the resulting functional regions reflect intrinsic properties of the
pretrained model rather than effects introduced by post-training.
The \emph{base} checkpoint consists of a vision encoder followed by a 28-layer transformer language backbone.
Interventions are applied to individual transformer layers indexed from early to
late depth.

\begin{figure*}[t]
  \centering
  \includegraphics[width=0.6\textwidth]{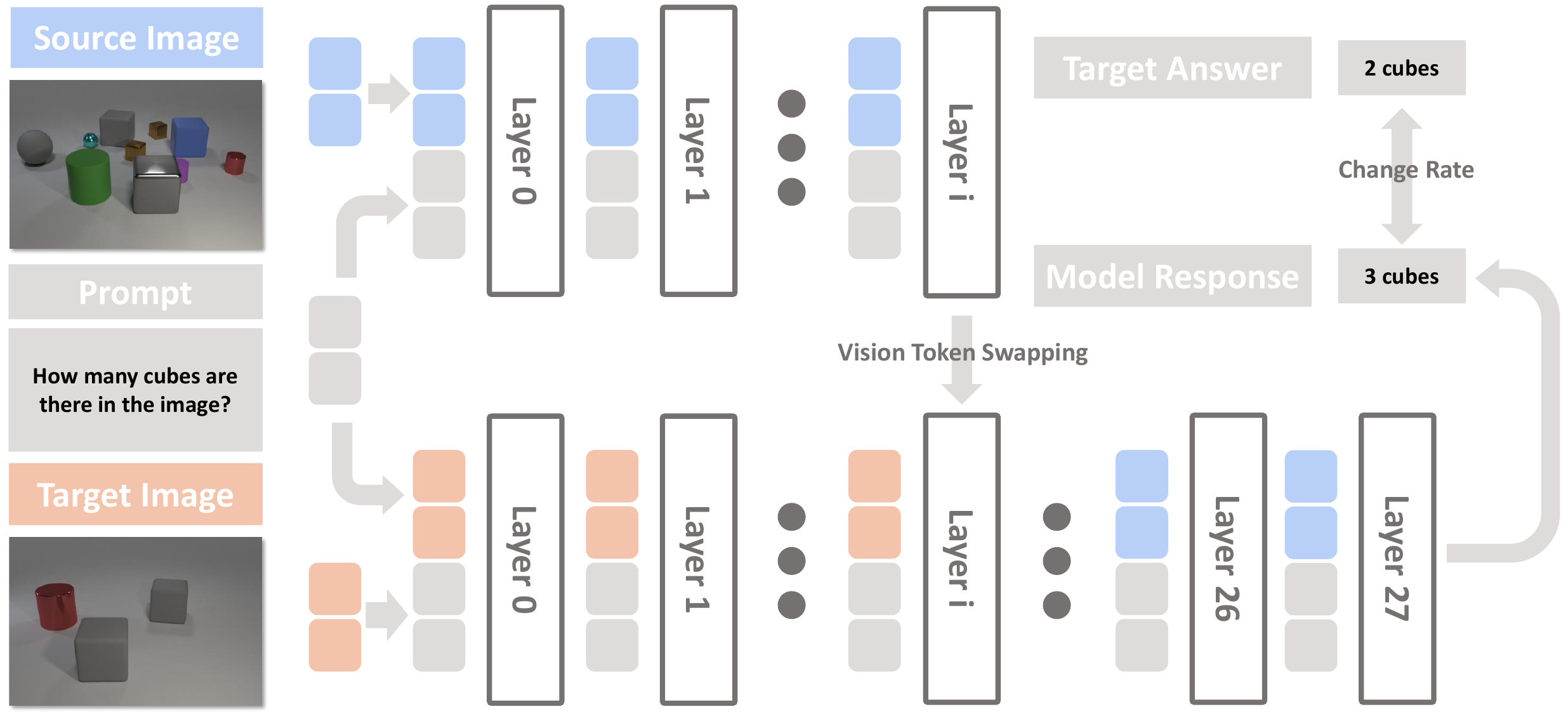}
\caption{\textbf{Illustration for vision token swapping to measure layer $i$'s contribution to model's response.}}
  \label{fig:appendix_vision_demo}
\end{figure*}

\subsection{Vision Functionality Localization}
\label{app:vision-localization}

\subsubsection{Layer-Wise Vision Token Swapping }

Vision functional localization is performed via \emph{vision token swapping}.
Given a pair of images, we replace the visual token sequence produced by one
image (source) with that of the other image (target) at a designated transformer
layer, while preserving all textual tokens and the decoding procedure.

Formally, let $\mathbf{v}^{(s)}$ and $\mathbf{v}^{(t)}$ denote the visual token
sequences corresponding to a source and target image.
At layer $k$, the hidden states associated with $\mathbf{v}^{(t)}$ are replaced
by those of $\mathbf{v}^{(s)}$, while all other hidden states remain unchanged.
The forward computation then proceeds normally.

This intervention preserves architectural structure and hidden-state scale,
ensuring that output changes reflect altered visual evidence rather than
numerical instability. A concrete demonstration could be found in Figure~\ref{fig:appendix_vision_demo}.

\subsubsection{Paired Image Dataset Construction}

To isolate individual visual functions, we construct paired image datasets in
which each image pair differs in exactly one visual attribute.
This enables functional attribution by ensuring that any output change can be
causally linked to the perturbed attribute.

We consider four visual functions:

\paragraph{Optical Character Recognition (OCR).}
OCR pairs consist of images containing different words rendered on visually
uniform blank backgrounds.
Words are sampled from a deduplicated arXiv corpus.
Queries ask the model to read and report the text content.
The change rate is defined as whether the generated text differs.

\paragraph{Object Counting.}
Counting pairs are adapted from the CLEVR~\citep{johnson2017clevr} dataset.
Each pair differs only in the number of instances of a target object category,
with object appearance and background held fixed.
The change rate is defined as whether the predicted count changes.

\paragraph{Object Grounding.}
Grounding pairs consist of identical objects placed at different random spatial
locations on otherwise clean backgrounds.
Queries request a bounding box for the target object.
Change is measured by whether the Intersection-over-Union (IoU) between the
predicted bounding box and the swapped ground-truth bounding box exceeds 0.5.

\paragraph{Object Recognition.}
Recognition pairs are drawn from COCO~\citep{lin2014microsoft} images containing a target object and
paired with visually blank canvases.
Queries ask whether the target object is present.
The change rate is defined as the proportion of ``No'' predictions following
token swapping.

\subsubsection{Change Rate Computation}

For each visual function and each transformer layer, we compute the
\emph{change rate} as the proportion of test instances for which the model’s
output differs between the original and swapped conditions.
Aggregating change rates across layers yields a layer-wise sensitivity profile,
where a higher change rate indicates stronger functional reliance on visual
information at that layer.

\subsubsection{Examples}

\begin{figure*}[t]
  \centering
  \includegraphics[width=1.0\textwidth]{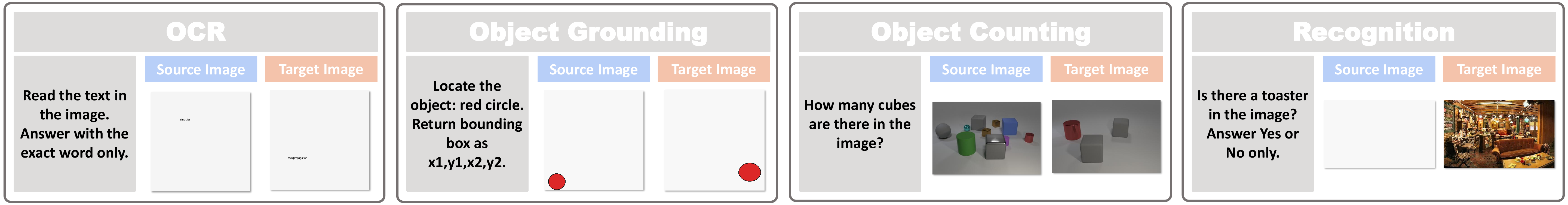}
\caption{\textbf{Examples of paired image datasets constructed to isolate visual functions.}}
  \label{fig:appendix_vision_example}
\end{figure*}

We provide an example for each task in Figure~\ref{fig:appendix_vision_example}.

\subsection{Reasoning Functionality Localization}

\label{app:reasoning-localization}

\subsubsection{Layer-wise Skipping}
To identify layers that are causally necessary for reasoning, we perform
\emph{layer-wise zero ablation} following prior analyses of mathematical
reasoning in transformer models.
For a given layer $k$, all trainable parameters in both the multi-head
self-attention and feed-forward submodules are set to zero, while residual
connections and normalization layers are preserved.

\begin{figure*}[t]
  \centering
  \includegraphics[width=0.8\textwidth]{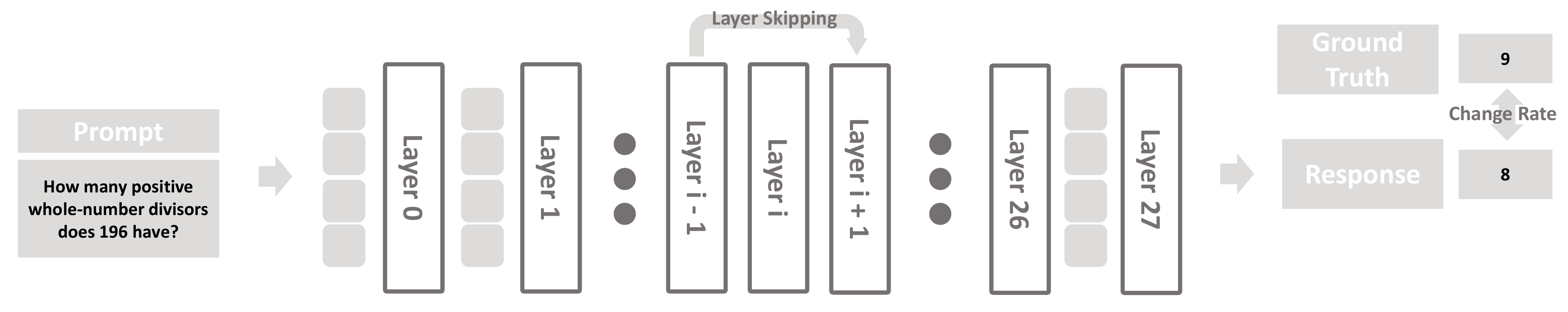}
\caption{\textbf{Illustration for layer skipping to measure layer $i$'s contribution to model's response.}}
  \label{fig:appendix_reason_demo}
\end{figure*}

This intervention reduces the ablated layer to an identity transformation while
leaving the model architecture and decoding process unchanged.
Each layer is ablated independently, and no gradients are computed during
evaluation. A concrete demonstration is provided in Figure~\ref{fig:appendix_reason_demo}.

\subsubsection{Reasoning Benchmarks}

Reasoning sensitivity is evaluated on reasoning-dominant, text-only benchmarks
that require multi-step inference.
We focus on \textbf{GSM8k}~\citep{cobbe2021training} and \textbf{MATH500}~\citep{lightman2023let}, a curated subset of competition-level mathematical
problems requiring structured derivation and symbolic manipulation.
These tasks do not depend on perceptual inputs and therefore isolate reasoning
computation.

\subsubsection{Layer-wise Contribution Profile}

Aggregating accuracy drops across evaluation samples yields a layer-wise
reasoning sensitivity profile.
Layers whose ablation causes substantial performance degradation are interpreted
as being functionally critical for reasoning, while layers with minor effects are
considered less essential.
This profile defines a \emph{reasoning functional region} rather than a single
exclusive locus of reasoning computation.

\subsection{Inference Settings}

For model inference, we summarize the experimental configurations used in the
functional localization studies in
Table~\ref{tab:app_location_inference_hyperparams}.
\begin{table*}[t]
    \rowcolors{2}{gray!11}{white}
    \centering
    \small

    \caption{
    \textbf{Inference hyperparameters for functional localization experiment.}
    }
    \label{tab:app_location_inference_hyperparams}

    \resizebox{0.25\columnwidth}{!}{
        \begin{tabular}{l|l}
            \thickhline
            \toprule
            \textbf{Parameter} & \textbf{Setting } \\
            \midrule

            max\_new\_tokens
                & 128  \\
            temperature
                & 0.0 \\
            do\_sample
                & Disabled \\
            \bottomrule
            \thickhline
        \end{tabular}
    }
\end{table*}

\subsection{Interpretation and Limitations}

The functional localization experiments in this appendix are designed as causal
probes rather than explanatory models of internal computation.
High sensitivity indicates necessity under intervention, not sufficiency or
exclusive responsibility.

Vision token swapping identifies where visual evidence is actively consumed,
while layer-wise zero ablation identifies layers whose computation is critical
for multi-step reasoning.
Together, these functional regions provide a unified reference frame for
interpreting decoupling analyses and RL-based post-training effects in the main
text.

\section{Model Merging}
\label{sec:appendix_model_merging}

This section details the implementation of the region-wise model merging
experiments used throughout the paper.

\paragraph{Layer region partition.}
Following the functional localization results in
Section~\ref{sec:location}, we partition the transformer backbone into three
contiguous regions of equal depth.
For the $28$-layer transformer used in our experiments, the partition is defined
as:
\begin{itemize}
    \item \textbf{\emph{\textcolor{earlyblue}{Early}}} region: layers $0$ to $9$,
    \item \textbf{\emph{\textcolor{midgrey}{Mid}}} region: layers $10$ to $18$,
    \item \textbf{\emph{\textcolor{latered}{Late}}} region: layers $19$ to $27$.
\end{itemize}
This 1/3-1/3-1/3 partition provides a coarse, depth-based functional reference
frame, enabling consistent comparison of region-wise effects across training
recipes.
We emphasize that these regions are not intended as precise architectural
boundaries, but as an abstraction for analyzing relative functional roles across
model depth.

\paragraph{Model merging procedure.}
Given an \textbf{IN} checkpoint and a \textbf{RL} checkpoint from the same training recipe, we construct hybrid models
by selectively transferring layers between checkpoints.
For each hybrid configuration, layers are sourced either from the IN or RL
checkpoint according to the specified region assignment, while all remaining
components are left unchanged.

For each selected layer, we directly copy the full parameter state of all layers
within that layer, including self-attention, feed-forward, and layer
normalization parameters.
No additional fine-tuning, re-normalization, or calibration is performed after
merging.

\section{Model Training with Region-wise Freezing}
\label{sec:appendix_model_training}

This section describes the training configuration used for the
\emph{region-wise parameter freezing} experiments in
Section~\ref{sec:necessity}.

\subsection{Implementation.}
All region-wise freezing experiments are implemented using the official
\texttt{OpenMMReasoner} training codebase\footnote{\url{https://github.com/EvolvingLMMs-Lab/OpenMMReasoner}}.
We follow the standard GRPO-based RL post-training pipeline provided by the
repository and modify it by freezing selected transformer layers during
training.
Specifically, parameters belonging to the frozen layers are excluded from
optimization, while all remaining parameters are updated normally.
No other changes are made to the training procedure.

\subsection{Hardware and training setup.}
All experiments are conducted using $2\times$ NVIDIA H200-SXM GPUs.
Each RL run is trained for $2000$ steps.
To ensure stable training under region-wise freezing and limited GPU memory, we
adopt reduced batch sizes and sequence lengths while keeping all other settings
identical across freezing conditions.

\subsection{Batching and rollout configuration.}
We use a per-device training batch size of $1$.
PPO minibatching is applied across GPUs, and for each prompt the model generates
multiple candidate responses to support reward-based optimization.
The full batching and rollout configuration used for all region-wise freezing
experiments is summarized in
Table~\ref{tab:app_freezing_training_hyperparams}.

\begin{table}[t]
    \rowcolors{2}{gray!11}{white}
    \centering
    \small

    \caption{
    \textbf{Training hyperparameters for region-wise freezing experiments.}
    }
    \label{tab:app_freezing_training_hyperparams}

    \resizebox{0.5\columnwidth}{!}{
        \begin{tabular}{l|l}
            \thickhline
            \toprule
            \textbf{Parameter} & \textbf{Setting} \\
            \midrule

            Training steps
                & 2000 \\

            GPUs
                & $2\times$ H200-SXM \\

            train\_batch\_size
                & 1 \\

            ppo\_mini\_batch\_size
                & 2 \\

            ppo\_micro\_batch\_size\_per\_gpu
                & 1 \\

            n\_rollout\_per\_prompt
                & 4 \\

            max\_response\_length
                & 2048 \\

            Frozen parameters
                & Selected transformer regions \\

            \bottomrule
            \thickhline
        \end{tabular}
    }
\end{table}

\end{document}